\documentclass[preprints,article,accept,pdftex,moreauthors]{Definitions/mdpi}

\usepackage{amssymb}
\usepackage{lipsum}
\usepackage{soul}
\usepackage{color}
\usepackage{booktabs}
\usepackage{amsmath}
\usepackage{graphicx}
\usepackage{placeins}
\usepackage{makecell}
\usepackage{tabu}
\usepackage{threeparttable}

\newcommand{\sens}{SCOUT }
\newcommand{\iv}{\textit{in-vivo }}
\newcommand{\ev}{\textit{ex-vivo }}
\newcommand{\co}{CO$_2$ }
\newcommand{\ch}{CH$_4$ }

\Title{SCOUT: Closed-Loop \textit{in-vivo} System for Continuous Methane Concentration Monitoring in Cattle}

\Author{Yuelin Deng $^{1}$, Hinayah Rojas de Oliveira $^{2}$, Richard M. Voyles $^{1,3}$ and Upinder Kaur $^{4,*}$}

\AuthorNames{Yuelin Deng, Hinayah Rojas de Oliveira, Richard M. Voyles and Upinder Kaur}

\address{%
$^{1}$ \quad Purdue Polytechnic, Purdue University, 401 N. Grant St., West Lafayette, Indiana 47907, USA; deng226@purdue.edu (Y.D.);\\
$^{2}$ \quad Department of Animal Sciences, Purdue University, 270 S Russell St, West Lafayette, Indiana 47907, USA; hrojasde@purdue.edu (H.R.d.O.)\\
$^{3}$ \quad University of Texas at Arlington Research Institute, The University of Texas at Arlington, 7300 Jack Newell Boulevard South, Fort Worth, Texas 76118, USA; richard.voyles@uta.edu (R.M.V.)\\
$^{4}$ \quad Department of Agricultural and Biological Engineering, Purdue University, 221 S University St, West Lafayette, Indiana 47907, USA; kauru@purdue.edu (U.K.)}

\corres{Correspondence: kauru@purdue.edu}

\abstract{Enteric methane measurement from ruminant livestock faces fundamental trade-offs between accuracy and operational feasibility. Existing methods quantify methane after eructation and atmospheric dilution, limiting temporal resolution and confounding biological signals with environmental variables. We present SCOUT (Smart Cannula-mounted Optical Unit for Trace-methane), the first autonomous system for continuous in-vivo monitoring of ruminal headspace methane concentrations. The system addresses a critical engineering barrier through closed-loop gas recirculation that maintains anaerobic ruminal conditions during persistent headspace sampling. SCOUT was deployed on cannulated Simmental heifers under contrasting dietary treatments. Headspace concentrations were 100 to 1000 times higher than concurrent ambient sniffer readings, providing substantially greater signal resolution for characterizing methane dynamics. High-frequency monitoring revealed behavior-production coupling previously inaccessible, including rapid concentration changes ($14.5 \pm 11.3k$ ppm) associated with postural transitions within 15-minute intervals. Cross-platform comparison with ambient sniffers showed scale-dependent correspondence between production and release measurements, with an optimal correlation (r = -0.564) at 40-minute averaging windows consistent with eructation cycles. These results demonstrate that the rumen headspace contains continuous, biologically interpretable methane signals that SCOUT can reliably access, establishing the measurement infrastructure necessary for developing concentration-to-flux models that would support precision phenotyping, emission proxy calibration, and mitigation strategy evaluation.}

\keyword{methane; sensing; edge sensing; automation; cattle}

\begin{document}

\fancyhf{}
\renewcommand{\headrulewidth}{0pt}
\renewcommand{\footrulewidth}{0pt}

\section{Introduction}
\label{introduction}
Enteric methane from ruminant livestock represents both an environmental challenge and a productivity opportunity in modern animal agriculture. Methane emissions account for approximately 14\% of global emissions, on a \co-equivalent basis,~\citep{Gerber2013, IPCC2021} while representing an energy loss of 8-12\% of gross feed intake~\citep{Hristov2018, knapp2014invited}. This dual impact on environmental sustainability and animal productivity motivates both climate-driven and productivity-driven mitigation efforts~\citep{smith2014agriculture, beauchemin2011mitigation}.

Progress in methane mitigation is constrained by measurement limitations across accuracy, temporal resolution, cost, and scalability. Precision livestock management requires measurements that are simultaneously accurate, temporally resolved, robust to real-world variability, affordable, and operationally scalable~\citep{garnsworthy2019comparison, hristov2015use}. Current methane measurement methodologies range from highly accurate but resource-intensive whole-animal respiration calorimetry to field-deployable but environmentally sensitive ambient sampling techniques, with no single approach satisfying these competing demands~\citep{hammond2015methanegreen, zhao2020review}. Table~\ref{tab:technology_comparison} summarizes the trade-offs among six individual-animal methods and one herd-level approach. Respiration chambers achieve measurement uncertainties below 2\%~\citep{Storm2012} but require infrastructure costs exceeding \$200,000 per unit and confine animals, limiting throughput to 30--100 animals annually~\citep{gardiner2015determination}. The SF\textsubscript{6} tracer technique enables field measurements but exhibits coefficients of variation of 15--30\%~\citep{grainger2007methane}. Automated visit-based systems such as GreenFeed (C-Lock Inc.) offer commercial phenotyping but depend on voluntary visit frequency and cost over \$100,000 per unit~\citep{hristov2015use, hammond2015methanegreen}. In-line sniffer systems provide high throughput at moderate cost (\$10,000--\$20,000) but measure concentration rather than flux, with measurement uncertainties commonly exceeding 25\% due to atmospheric dilution, head positioning, and environmental conditions~\citep{garnsworthy2012variation, lassen2012accuracy, wu2018uncertainty}. Manual handheld laser detectors enable rapid screening but show only weak to moderate correlations (r = 0.3--0.6) with chamber references~\citep{chagunda2013measurement, ricci2014evaluation}.

All established individual-animal methods share a fundamental characteristic that they measure methane concentrations after eructation and atmospheric dilution, at the point of release rather than at the site of generation in the rumen. Environmental factors such as wind, temperature, humidity, and animal positioning introduce variability that obscures biological signals. For example, the coefficient of determination between known methane release rates and sniffer measurements decreases from 0.97 in controlled laboratory settings to 0.37 in commercial barn environments~\citep{wu2018uncertainty}. This environmental confounding limits both measurement precision and the ability to validate mechanistic models of ruminal methanogenesis~\citep{dijkstra1992simulation, Hristov2018}, which predict production kinetics but have lacked continuous production-level data for direct validation.

Direct measurement of ruminal headspace methane concentrations would provide access to fermentation-linked dynamics before atmospheric mixing, offering substantially higher signal-to-noise ratios for characterizing methane dynamics. However, continuous \iv sensing has remained inaccessible due to strict engineering constraints. Methane monitoring requires persistent gas-phase sampling while maintaining the strict anaerobic conditions necessary for normal rumen function. Existing indwelling rumen sensors, such as boluses developed for drug delivery and biomarker monitoring (temperature, pH)~\citep{kaur2023invited,han2022invited}, settle in the liquid phase at the rumen bottom and cannot access the gas headspace where methane accumulates prior to eructation. Previous attempts at ruminal gas sampling were limited to manual single-point collection or required continuous venting that introduces atmospheric oxygen and compromises methanogenic archaea populations~\citep{Storm2012}.

\begin{figure}[ht]
    \centering
    \includegraphics[width=\linewidth, trim=0in 0.1in 4in 0.1in, clip]{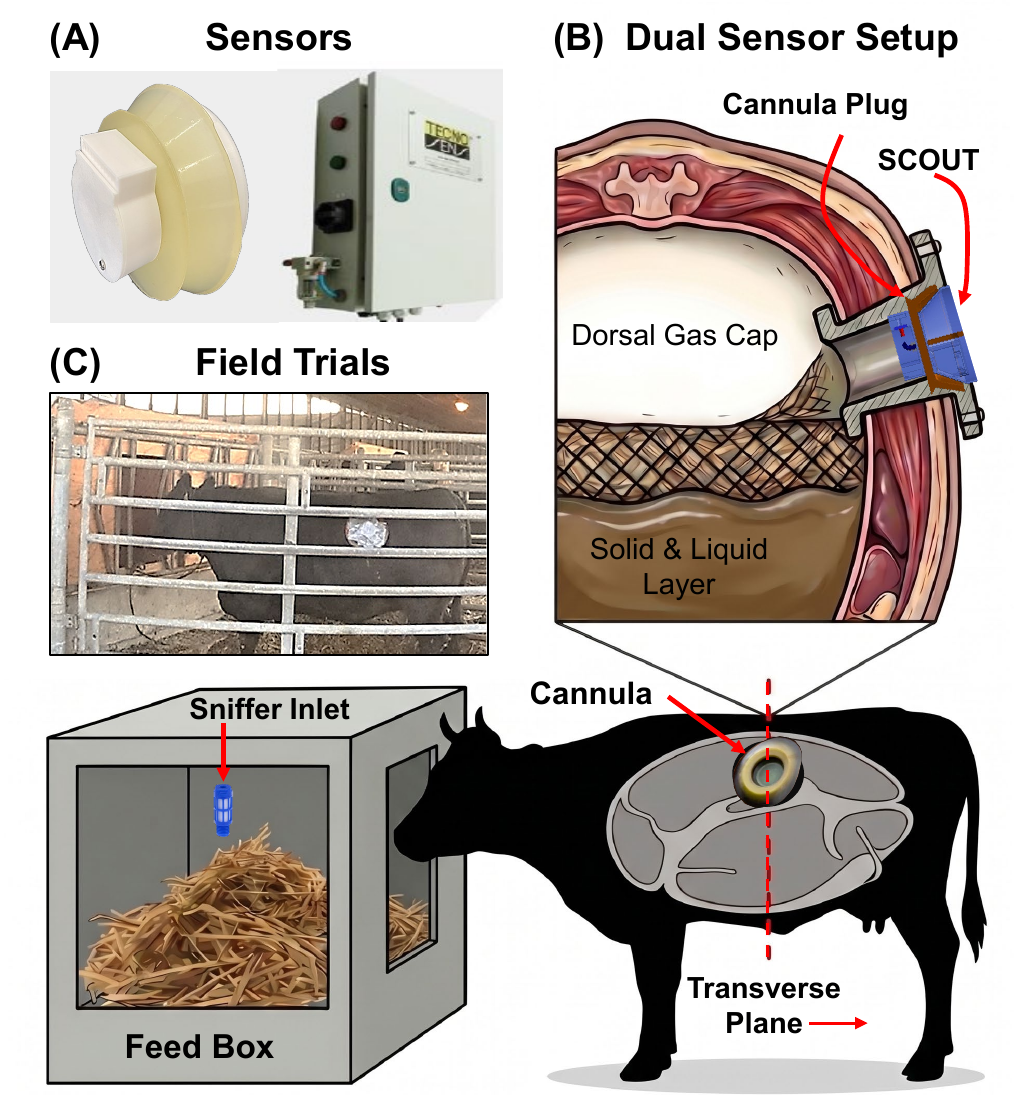}
    \caption{Overview of the SCOUT experimental framework.
        (A) SCOUT sensor device (left) and MooLogger sniffer system (right).
        (B) Illustrated experimental configuration with sniffer inlet in the feed box and SCOUT on the cannula plug. Inset: transverse cross-section showing the \textit{in-vivo} chamber in the dorsal gas cap and \textit{ex-vivo} housing externally.
        (C) Field trial deployment in a cattle barn.}
    \label{fig:farm_setup}
\end{figure}

We address this measurement gap with SCOUT (Smart Cannula-mounted Optical Unit for Trace-methane), an intraruminal monitoring system enabling continuous characterization of methane production dynamics in the rumen headspace. SCOUT uses a cannula-mounted interface, building on our earlier work~\citep{kaur2022casper}, and integrates a closed-loop gas-recirculation mechanism that enables continuous headspace sampling while preserving anaerobic conditions. Figure~\ref{fig:farm_setup} provides an overview of the experimental framework, including sensor hardware; a cross-sectional view of the cannula-mounted installation showing how the \textit{in-vivo} chamber accesses the rumen headspace, while the \textit{ex-vivo} chamber houses the electronics externally; and the field trial deployment. We benchmark the operational performance of SCOUT in field trials with cannulated heifers under contrasting dietary treatments and compare its measurements with concurrent ambient sniffer readings to assess cross-platform correspondence.

This study makes four primary contributions to the field of ruminal methane measurement. First, it presents an intraruminal methane sensing architecture that achieves autonomous, continuous operation through closed-loop gas recirculation while preserving the anaerobic conditions required for normal rumen function. Second, it provides quantitative characterisation of behavioral-concentration coupling at sub-minute temporal resolution, demonstrating dynamic patterns previously inaccessible through ambient measurement methods. Third, it establishes cross-platform correspondence between \iv rumen headspace concentrations and ambient sniffer readings at physiologically relevant timescales, confirming that SCOUT captures genuine eructation biology rather than measurement artifact. Fourth, it produces a high-resolution temporal dataset characterising diurnal patterns, individual variation, and diet-induced differences in ruminal methane dynamics. Overall, this lays the foundation for highly precise and real-time concentration-to-flux models, quantifying methane dynamics for cattle for applications in precision phenotyping and evaluation of mitigation strategies.

This paper is organized as follows. Section~\ref{sec:sensor} describes the SCOUT system design, including the closed-loop recirculation architecture, sensing technology, and control system. Section~\ref{sec:data} details the data processing and calibration frameworks for both SCOUT and the benchmarking sniffer system. Section~\ref{sec:method} presents the experimental methodology, including animal subjects, dietary treatments, and deployment protocols. Section~\ref{sec:results} reports results on system performance, cross-platform correspondence, and behavioral correlations. Section~\ref{sec:discussion} discusses the implications and limitations of the work, including the cannula requirement, sensor saturation, sample size, and the pathway toward emission quantification.

\begin{table*}[htbp]
    \centering
    \caption{Comparison of methane measurement methods for ruminant livestock}
    \label{tab:technology_comparison}
    \resizebox{\textwidth}{!}{%
        \begin{tabular}{lcccccccc}
            \toprule
            \textbf{Method}         & \textbf{Deployment} & \textbf{Cost}      & \textbf{Labor}     & \textbf{Environmental}   & \textbf{Data}         & \textbf{Temporal}       & \textbf{Animal}       \\
                                    & \textbf{Scale}      & \textbf{(Capital)} & \textbf{Intensity} & \textbf{Sensitivity}$^a$ & \textbf{Coverage}$^b$ & \textbf{Resolution}$^c$ & \textbf{Behavior}$^d$ \\
            \midrule
            Respiration Chambers    & Individual          & Very High          & High               & None                     & High                  & High                    & Restricted            \\
            \addlinespace
            SF$_6$ Tracer Technique & Group               & High               & High               & Medium                   & Medium                & Low                     & Normal                \\
            \addlinespace
            GreenFeed System        & Group               & High               & Low                & Low                      & Medium$^e$            & Medium$^e$              & Modified              \\
            \addlinespace
            Head-box/Sniffer        & Herd                & Medium             & Low                & Medium                   & Low                   & Medium                  & Normal                \\
            \addlinespace
            Manual Sampling         & Individual          & Low                & High               & High                     & Low                   & Varies                  & Disrupted             \\
            \addlinespace
            Micrometeorological     & Herd                & Medium             & Low                & High                     & N/A                   & Low                     & Normal                \\
            \addlinespace
            \textbf{SCOUT (Ours)}   & \textbf{Individual} & \textbf{Low}       & \textbf{Low}       & \textbf{None}            & \textbf{High}         & \textbf{High}           & \textbf{Normal}       \\
            \bottomrule
        \end{tabular}%
    }
    \begin{tablenotes}
        \small
        \item $^a$\textbf{Environmental Sensitivity}: Degree to which external conditions (wind, temperature, humidity) affect measurement accuracy. None = measurements unaffected by environmental conditions (controlled environment or isolated from atmosphere); Low = minimal impact requiring minor corrections; Medium = moderate impact requiring calibration adjustments; High = substantial weather-dependent variability compromising data quality
        \item $^b$\textbf{Data Coverage}: Proportion of an individual animal's methane production/release captured by the measurement approach on average over 24 hours. Low = sporadic sampling; Medium = regular but incomplete sampling; High = near-complete emission profile. Note that SCOUT measures production in the rumen headspace before release, while other methods measure post-eructation release.
        \item $^c$\textbf{Temporal Resolution}: Degree to which a measurement approach captures short-term fluctuations in methane dynamics, reflecting both sampling frequency and the continuity of individual-level observation. High = continuous or near-continuous tracking; Medium = repeated discrete events allowing partial reconstruction of temporal patterns; Low = sparse or discontinuous sampling representing time-averaged values.
        \item $^d$\textbf{Animal Behavior}: Impact on natural feeding, rumination, and movement patterns. Normal = unrestricted natural behavior; Modified = requires behavioral adaptation (e.g., visiting feed stations); Restricted = confined movement in chambers; Disrupted = intermittent human interaction affecting behavior
        \item $^e$ Although the GreenFeed system can achieve high temporal coverage and resolution, this level of performance relies on consistent animal visits across the 24-hour cycle and the application of the manufacturer's (or research protocol's) data-processing algorithms (e.g., cleaning, weight-averaging, time-bin interpolation, visit-duration filters). Without careful scheduling of visits and disciplined post-processing, the dataset will more likely reflect population-level averages rather than detailed, continuous individual-animal emission dynamics.
    \end{tablenotes}
\end{table*}

\section{SCOUT Sensor Design and Configuration} \label{sec:sensor}
\begin{figure*}[ht]
  \centering
  \includegraphics[width=\linewidth, trim=0in 0.2in 0in 0.2in, clip]{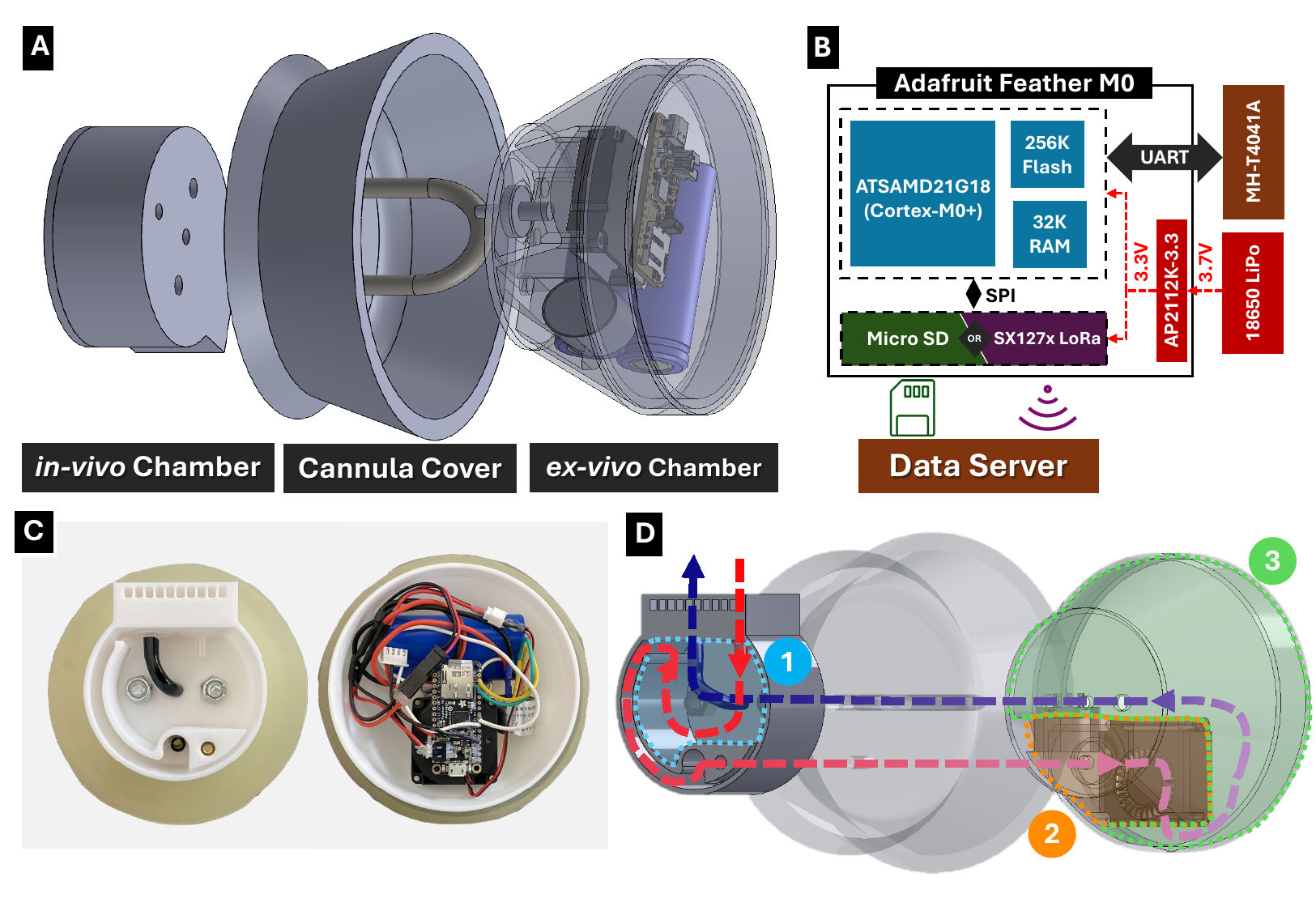}
  \caption{
    Overview of \sens.
    (A) Exploded CAD view showing the \iv water-trap chamber, cannula cover, and \ev sensor housing.
    (B) Circuit layout of the embedded microcontroller, MH-T4041A methane sensor, and data storage (LoRa/microSD) links.
    (C) Prototype photographs: rumen-side intake (left) and electronics/battery pack (right).
    (D) Annotated airflow diagram of the closed-loop gas circuit; numerals 1–3 correspond to the water trap, plenum, and stabilization/sensor chamber.
  }
  \label{fig:scout_design}
\end{figure*}

The SCOUT system addresses the fundamental engineering challenges of \iv rumen monitoring through a modular architecture prioritizing environmental compatibility, measurement accuracy, and autonomous operation (Figure~\ref{fig:scout_design}). Our design integrates three primary innovations: cannula-mounted gas sampling, NDIR-based sensor module with detection and control electronics, and closed-loop recirculation maintaining anaerobic conditions during continuous operation.

\subsection{Design Requirements and Environmental Constraints}

The rumen presents extreme conditions for sensor deployment: near-anaerobic atmosphere (O\textsubscript{2} $<$0.1\%), temperatures of $38$--$42^{\circ}$C, relative humidity approaching 100\%, pH of $5.8$--$6.8$, and cyclic pressure fluctuations of $\pm0.2$~kPa from reticulorumen contractions~\citep{martin2010methane, wang2019effects}. The most critical constraint is preserving anaerobiosis, as oxygen introduction disrupts \textit{methanogenic archaea} populations and alters fermentation pathways~\citep{Storm2012}. All materials must also resist the corrosive effects of rumen fluid~\citep{pitt1996prediction}. SCOUT addresses these constraints through a cannula-mounted interface, a surgically installed access port widely used in ruminant research, avoiding additional surgical procedures while providing consistent access to rumen headspace gases.

\subsection{Mechanical Design and Materials Selection}
The SCOUT system has two primary modules mounted on either side of a standard 4-inch rumen cannula: an \textit{in-vivo} chamber housed within the rumen headspace for gas sampling, and an \textit{ex-vivo} chamber positioned externally, containing the sensor electronics and control hardware. The mechanical layout is illustrated in the exploded CAD rendering in Figure~\ref{fig:scout_design}A, while Figure~\ref{fig:scout_design}C shows the fully assembled prototype, with the rumen-facing intake on the left and the external electronics enclosure on the right.

Material selection prioritized chemical resistance, thermal stability, and biocompatibility. The primary housing is fabricated from medical-grade PETG (polyethylene terephthalate glycol), chosen for its robustness under thermal cycling and resistance to organic acids commonly found in the rumen. Gas transfer tubing uses medical-grade silicone (inner diameter 3.2 mm, wall thickness 1.6 mm), selected for its low oxygen permeability, flexibility, and proven biocompatibility in long-term biomedical applications.

The \textit{in-vivo} water-trap chamber is positioned within the cannula opening and functions as a buffer to prevent liquid intrusion into the sampling pathway. It incorporates a passive liquid separation mechanism based on gravitational settling and surface tension, with an internal baffle system that minimizes carryover without impeding gas flow. A replaceable $0.2 \mu$m PTFE (Polytetrafluoroethylene) membrane filter at the gas inlet provides critical protection against particulate contamination, particularly important during in-rumen operation, where fine feed particles can damage sensitive optical components downstream.

\subsection{Closed-Loop Gas Recirculation System}
The maintenance of anaerobic conditions during continuous gas sampling represents the primary technical innovation of the SCOUT system. Traditional sampling approaches either require continuous venting of rumen gases to the atmosphere, leading to oxygen backflow, or employ single-pass sampling, which restricts sufficient gas exchange for representative measurements. Our SCOUT system addresses this challenge through a fully closed-loop recirculation mechanism that continuously circulates rumen gases through the sensor chamber while maintaining strict isolation from atmospheric oxygen.

The recirculation system employs a low-power radial fan (40×40×10 mm) operating at reduced voltage to generate a controlled pressure differential of approximately 0.25 mbar across the gas circuit. This pressure differential overcomes flow resistance through the tubing and sensor chamber while remaining below levels that could disrupt normal rumen pressure dynamics. Natural rumen contractions also contribute beneficially to gas circulation, generating periodic pressure pulses that enhance mixing and ensure representative sampling across the rumen headspace.  Gas flow rates typically range from $0.5-1.2 L/min$ depending on rumen pressure conditions and fan speed settings.

The gas circulation pathway consists of three distinct chambers, each optimized for specific functional requirements. Chamber 1 serves as the primary water trap and pressure buffer, with a volume of $43$ mL and internal geometry designed to promote liquid separation through gravitational settling. Chamber 2 functions as a negative-pressure plenum with a volume of $25$ mL, providing surge capacity to smooth pressure fluctuations from rumen contractions. Chamber 3 represents the primary sensor chamber with a volume of $209$ mL, designed to provide adequate residence time for optical measurements while minimizing dead volume effects. Figure~\ref{fig:scout_design}D annotates the three chambers (1 – water trap, 2 – plenum, 3 – sensor chamber) and the bidirectional flow paths that realise this closed loop.

\subsection{Methane Sensing Technology}
The methane detection system centers on an MH-T4041A infrared absorption sensor selected for its broad detection range (0-50,000 ppm), high resolution (100 ppm), and robust performance under high-humidity conditions up to 95\% relative humidity. The sensor employs non-dispersive infrared (NDIR) technology, measuring methane concentration through selective absorption of infrared radiation at the characteristic $3.3\mu m$ wavelength. The sensor has low power consumption (\textless 30~mA operating current), making it well-suited for battery-powered autonomous deployment. The sensor housing maintains IP65 environmental protection while providing optical access through anti-reflective coated windows that minimize signal loss and prevent condensation buildup.

\subsection{Control System and Data Acquisition}
The control system employs an Adafruit Feather M0 Adalogger microcontroller featuring an ARM Cortex-M0+ processor, integrated microSD card interface for local data storage, and low-power design optimized for autonomous field deployment. The microcontroller manages sensor communication via UART protocol, implements data logging routines, and controls the gas circulation fan through pulse-width modulation. Figure~\ref{fig:scout_design}B summarises the hardware stack and data pathways, including optional LoRa telemetry and on-board microSD logging.

Data acquisition occurs at a sampling rate of 0.1 Hz, selected based on the characteristic time scales of rumen methane production and the need to balance temporal resolution with power consumption constraints. Previous research indicates that significant methane production variations occur on timescales of minutes to hours, making 0.1 Hz sampling sufficient to capture relevant biological dynamics while avoiding unnecessary power overhead. Each data record includes timestamps, methane concentration, internal temperature, and system status information, stored in CSV format for compatibility with standard analysis software.

Power is supplied by dual 2000~mAh lithium-ion batteries in parallel (4000~mAh total, 3.7~V nominal) with low-dropout regulation to 3.3~V. Under typical operating conditions (mean power draw: 444~mW during active circulation, 40~mW idle), the system achieves continuous operation exceeding 24~hours.

\section{Data Processing and Calibration Frameworks}
\label{sec:data}
In this section, we detail the data processing and calibration steps for the \sens sensor and the MooLogger sniffers (Tecnosense, Italy) used for cross-platform comparison purposes.

\subsection{Signal Preprocessing and Quality Control}
\label{sec:signal_preprocess}

\subsubsection{Sensor Response Characterization: }
Laboratory step-response characterization measured SCOUT's intrinsic time constant at $\tau_{SCOUT} < 1$~s, though the 0.1~Hz field sampling rate is the practical limiting factor for temporal resolution. The MooLogger sniffer exhibited a longer time constant of $\tau_{MooLogger} = 2.754$~s, attributable to gas transport delays through its sampling pathway. In field deployments, additional factors such as wind speed, cow head orientation, and animal movement further influence effective equilibration time for the sniffer system.

\subsubsection{Daily Drift Verification: }
Both sensors were factory-calibrated against traceable gas standards before deployment. The field procedures described here serve only to monitor and correct for daily drift, not to recalibrate the instruments. Daily zero-offset verification used ambient air sampling (1.8--2.1~ppm, consistent with global background methane) to confirm absence of significant drift, with weekly checks demonstrating drift rates below 0.5\% per week. Sensor initialization data (2--3~minutes post-power-up) and occasional NaN values ($<$0.1\% of measurements) were removed. Temperature compensation algorithms integrated within the sensor hardware automatically corrected for thermal drift.

\subsubsection{Temporal Synchronization: }
The \sens system initializes with network-synchronized UTC timestamps but relies on internal crystal oscillator timing during autonomous operation. Cumulative timing drift ranged from $-30$ to $+45$~seconds over 24-hour periods and was corrected via linear interpolation using manual deployment records, ensuring temporal alignment between \sens measurements and concurrent sniffer and video data streams.

\subsection{Sniffer Data Processing}

Sniffer data collected at 1~Hz included \ch and \co concentrations (mg/m\textsuperscript{3}), volumetric flow rate, ambient temperature, and barometric pressure. Systematic artifacts from mandatory pump restart and water purging events were automatically identified via flow rate monitoring; conservative exclusion windows (2~s before to 40~s after each event) and a minimum flow rate threshold of 0.75~L/min were applied. Methane concentrations were converted from mass-based to volumetric units (ppm) using the ideal gas law corrected for real-time temperature and pressure:
\begin{equation}
    ppm_v = \frac{(mg/m^3 \times R \times (T_C + 273.15))}{(M_{CH_4} \times (P_{mbar} \times 100)) \times 1000}
\end{equation}
where $R = 8.314462$~J/(mol$\cdot$K), $M_{CH_4} = 16.04$~g/mol, $P_{mbar}$ is barometric pressure in millibars, and $T_C$ is ambient temperature in Celsius.

\subsection{Signal Filtering}

\begin{figure}
    \centering
    \includegraphics[width=\linewidth]{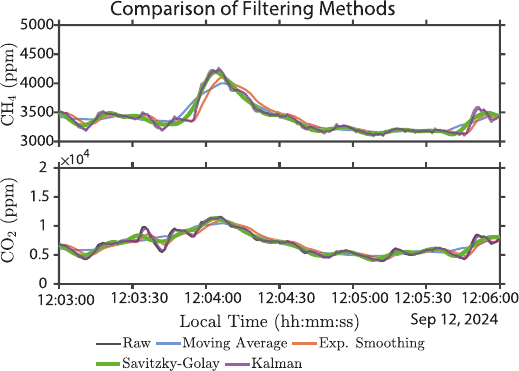}
    \caption{Comparison of digital filters for sniffer signal processing. Savitzky-Golay filter (solid green) best preserves transient emission peaks while maintaining low phase lag, compared to moving average (MA), exponential smoothing (ES), and Kalman filtering (KF). Top: \ch concentration. Bottom: \co concentration.}
    \label{fig:sniffer-filtering}
\end{figure}

Digital filtering of sniffer signals required suppressing ambient noise while preserving the brief, high-amplitude concentration spikes characteristic of discrete eructation events. Four candidate filters were evaluated on the CH\textsubscript{4} and CO\textsubscript{2} time series: moving average (21-sample window), exponential smoothing ($\alpha = 0.3$), Savitzky-Golay (21-sample window, third-order polynomial), and a constant-velocity Kalman filter. As shown in Figure~\ref{fig:sniffer-filtering}, the Savitzky-Golay filter best retained peak amplitude and timing while maintaining low phase lag, critical for detecting discrete eructation events. Moving average and exponential smoothing attenuated peaks and introduced phase delay, while the Kalman filter produced artifacts in high-gradient regions. Savitzky-Golay filtering was therefore adopted for all subsequent analyses.

\subsection{Ambient Baseline Correction and Normalization}
\label{sec:baseline}

Accurate quantification of animal-specific methane signals requires separating them from background ambient levels. Although the sniffer system was installed in individual feeding hoods, each hood remained open to the barn environment on one side, meaning measured concentrations reflected both the experimental subject's emissions and drifting ambient methane from neighboring pens. A two-stage baseline correction procedure was developed using CO\textsubscript{2} measurements as proxies for animal presence.

In the first stage, animal activity periods were identified using CO\textsubscript{2} concentration thresholds: absolute concentrations exceeding 350~ppm above daily median values, or first-order differences exceeding 175~ppm, indicating rapid changes during approach or departure events. Data outside these activity windows underwent secondary screening using a smoothed CO\textsubscript{2} baseline (2000-point moving average) with more conservative thresholds (250~ppm absolute, 125~ppm differential) to capture subtle activity signatures missed in the initial pass.

\begin{figure}
    \centering
    \includegraphics[width=\linewidth]{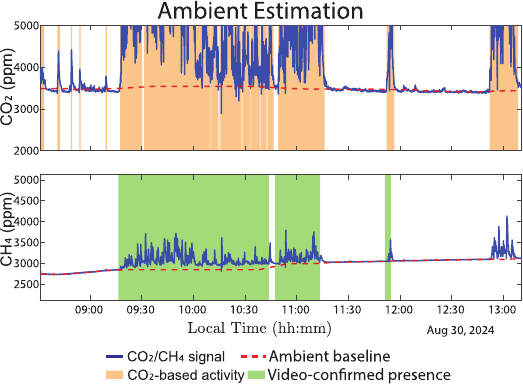}
    \caption{Two-stage baseline correction for ambient methane estimation. (a) \ch signal with dynamic ambient baseline (red dashed) and automatically detected activity windows (orange shading). (b) \ch signal with corresponding ambient correction. Green shading indicates video-confirmed cow presence, validating the CO$_{\text{2}}$-based detection algorithm.}
    \label{fig:sniffer-ambient}
\end{figure}

In the second stage, methane concentrations during confirmed no-animal periods were smoothed using 1000-point moving averages to generate a time-varying ambient baseline accounting for diurnal and facility-wide emission patterns. Subtracting this background profile from the filtered signal yielded a normalized methane trace isolating animal-specific events. Figure~\ref{fig:sniffer-ambient} illustrates this procedure; temporal alignment between CO\textsubscript{2}-based detection windows and video-confirmed cow presence validates the effectiveness of the approach. Threshold values were empirically optimized using video-confirmed presence periods to maximize detection sensitivity while minimizing false positives.

\subsection{Cross Platform Statistical Analysis}
\label{sec:calibration_method}

Cross-platform comparison required specialized statistical approaches due to fundamental differences between SCOUT and sniffer measurement contexts. SCOUT provides continuous monitoring of methane concentrations within the sealed ruminal environment, whereas sniffer systems detect atmospheric methane only when animals are positioned under feeding hoods and actively eructating. This discontinuity restricts meaningful comparisons to discrete eructation events where both systems detect the same biological process. This comparison does not provide quantification of absolute emission rates, which requires future validation against respiration chambers.

We identified candidate eructation events using \sens data, focusing on abrupt concentration drops consistent with ruminal gas expulsion during rumen contractions (Figure~\ref{fig:cross-trend}). Events were filtered for periods with video-confirmed cow presence under the feeding hood and concurrent non-zero sniffer methane readings. Scale-dependent correlation analysis employed sliding windows across temporal scales from 5--40~minutes (5-minute increments, 1-minute advancement). The negative correlation reflects the expected physiological coupling: ruminal headspace concentrations decrease during eructation as gas is expelled, producing corresponding concentration increases at the ambient sniffer.

Temporal synchronization accuracy was verified to within $\pm$15~seconds. We applied detrending (MATLAB's \texttt{detrend}) to remove long-term trends, and used signed correlation coefficients ($\text{R} = \text{sign}(\text{slope}) \times \sqrt{R^2}$) to preserve directional information crucial for interpreting the inverse production--release relationship. Multiple comparison correction employed Benjamini-Hochberg false discovery rate (BH-FDR) adjustment (q $<$ 0.05), and temporal autocorrelation was addressed using AR(1) effective sample size correction: $n_{eff} = n \times (1 - \rho_{x1} \times \rho_{y1}) / (1 + \rho_{x1} \times \rho_{y1})$.

\section{Experimental Methodology}
\label{sec:method}
The experimental framework assessed SCOUT system performance under controlled field conditions through three complementary approaches: (1) performance characterization under contrasting dietary treatments, (2) cross-platform comparison with an ambient sniffer system (MooLogger, Tecnosense), and (3) behavioral correlation analysis linking concentration dynamics to animal activity.

\subsection{Experimental Facility and Animal Subjects}
We conducted trials at the Purdue University Animal Sciences Research and Education Center between June 21 and September 19, 2024, using protocols approved by the Institutional Animal Care and Use Committee (IACUC protocol 0324002489). Four cannulated Simmental heifers served as experimental subjects: Bonita, Brittany, Gracie, and Bianca (approximately 20 months of age and ~1350 lb each). All animals had been previously cannulated with standard 4-inch rumen cannulas and acclimated to research handling for at least four weeks. Data from all four animals informed iterative system development; the quantitative results reported in Section~\ref{sec:results} derive from the final system configuration deployed on Bonita (high-grain diet) and Brittany (high-forage diet). Animals were housed in individual pens with feed bunks configured to accommodate gas sampling equipment. Feed was delivered twice daily at 0600 and 1500~hours, provided \textit{ad libitum}, with water continuously available through automatic systems positioned outside gas sampling areas.

Two contrasting dietary treatments were implemented to maximize differences in ruminal fermentation patterns and methane production rates (Table~\ref{tab:diet-characteristics}). High-grain diets typically reduce methane production by 15-30\% compared to forage-based systems due to shifts in volatile fatty acid production and reduced fiber fermentation~\citep{beauchemin2011mitigation}. Both diets meet or exceed NRC nutrient requirements for maintenance in mature beef cattle.

\begin{table}[ht]
    \centering
    \small
    \caption{Animal Diet Composition}
    \begin{tabular}{lcc}
        \toprule
        \textbf{Ingredient}         & \textbf{High-Forage Diet} & \textbf{High-Grain Diet} \\
                                    & \textbf{\% dry matter}    & \textbf{\% dry matter}   \\
        \midrule
        Dry rolled corn             & 30.0                      & 68.3                     \\
        \makecell[l]{Dried distillers grains                                               \\with solubles} & 19.0 & -- \\
        Bagged triticale hay        & 23.0                      & --                       \\
        Corn silage                 & 22.0                      & --                       \\
        Grass silage                & --                        & 10.0                     \\
        Soybean hulls               & --                        & 6.7                      \\
        Soybean meal                & --                        & 9.5                      \\
        Supplement                  & 6.0                       & 5.5                      \\
        \midrule
        \textbf{Total}              & \textbf{100.0}            & \textbf{100.0}           \\
        \midrule
        \textbf{Forage:Concentrate} & \textbf{45:55}            & \textbf{10:90}           \\
        \bottomrule
    \end{tabular}
    \label{tab:diet-characteristics}
\end{table}

\subsection{Instrumentation Setup and Deployment Protocol}

The experimental setup incorporated two measurement systems for cross-platform comparison. SCOUT sensors were mounted directly on rumen cannulas for continuous \iv monitoring of ruminal headspace methane concentrations, as illustrated in Figure~\ref{fig:farm_setup}(B). Concurrent ambient measurements employed MooLogger sniffer systems (Tecnosense, Italy) positioned within partially enclosed feeding chambers.

The sniffer intake in each animal's feed bunk was housed in a three-sided enclosure that formed a quasi-static sampling volume when the animal inserted its head to feed. Gas sampling lines were positioned approximately 10~cm above the feed surface and 15~cm from the chamber back wall, with sample flow rates maintained at 1.1~L/min through calibrated mass flow controllers. The enclosure design balanced gas concentration through natural sealing with the animal's head while maintaining adequate ventilation to prevent CO\textsubscript{2} buildup that could deter feeding.

Video monitoring (Sony FDR-X3000, 1920$\times$1080 at 30~fps) captured continuous behavioral data with timestamps synchronized to the same UTC reference used for sensor logging. Manual annotation identified key behavioral events, including feeding, postural changes (standing, lying, sitting), and water consumption.

\subsection{Data Collection Protocol}

Following iterative hardware refinements informed by initial deployments with all four animals, the finalized protocol was implemented with Bonita and Brittany. SCOUT sensors were deployed daily at 0800~hours for continuous 24-hour monitoring periods, following completion of morning feeding routines. Deployment procedures included verification of cannula seal integrity, gas flow pathway continuity, and data logging functionality (real-time monitoring for the first 10~minutes). Video cues and CO\textsubscript{2} concentration increases were combined to establish animal presence in the sniffer setup, with manual annotation of behavioral events including feeding initiation and termination, postural changes, and water consumption.

\section{Experimental Results}
\label{sec:results}

The results presented here derive from data collected using the final system configuration deployed on Bonita and Brittany, ensuring consistency in hardware specifications, data processing protocols, and measurement conditions.

\subsection{Data Collection Summary}
Data collection yielded 200 hours of simultaneous measurements across both animals and sensor systems. After applying preprocessing protocols (Section~\ref{sec:baseline}), effective data retention rates exceeded 82\% for SCOUT measurements and 78\% for sniffer measurements. Video-confirmed animal activity accounted for 17.4\% (Bonita) and 16.9\% (Brittany) of total monitoring time.

\subsection{SCOUT System Performance Characteristics}
\label{sec:scout_preformance}
\sens measurements revealed biological differences between dietary treatments, demonstrating sensor sensitivity to detect diet-induced changes in ruminal fermentation patterns. Sensor saturation events, defined as periods when measured concentrations exceeded the 50{,}000 ppm upper detection limit, accounted for 35\% and 41\% of total measurements for Bonita and Brittany, respectively. These saturation periods typically occurred 2--4 hours after major feeding events, consistent with established patterns of ruminal fermentation kinetics. This saturation during peak post-feeding fermentation constrains quantitative analysis when fermentation activity is highest; future sensor iterations will require an expanded measurement range.

Rapid concentration drops indicative of eructation events represented approximately 7.9\% and 6.4\% of measurements for Bonita and Brittany, providing markers for cross-platform comparison. Low values (\textless 1,000 ppm) comprised 20.8\% and 15.3\% of measurements, likely resulting from feed residue blocking the intake or condensate water accumulating in the pipeline. After excluding these compromised measurements, the effective data retention rate of 82\% represents periods of reliable concentration measurements suitable for pattern analysis. Table~\ref{tab:scout-signal-processing} summarizes key SCOUT signal characteristics.

\begin{table*}[htbp]
    \centering
    \small
    \caption{SCOUT Signal Characteristics Summary}
    \begin{tabular}{lccccccc}
        \toprule
        \textbf{Cow ID} & \textbf{\# Samples} & \textbf{Q$_{25}$} & \textbf{Q$_{50}$} & \textbf{Q$_{75}$} & \textbf{Q$_{90}$} & \textbf{\% Saturation} \\
                        &                     & \textbf{(ppm)}    & \textbf{(ppm)}    & \textbf{(ppm)}    & \textbf{(ppm)}    &                        \\
        \midrule
        Bonita          & 25,409              & 2,500             & 2,500             & 50,000            & 50,000            & 34.5704                \\
        Brittany        & 26,934              & 4,300             & 31,900            & 50,000            & 50,000            & 40.6067                \\
        \bottomrule
    \end{tabular}
    \label{tab:scout-signal-processing}
\end{table*}

\subsection{Sniffer Performance Characteristics}

During active feeding periods, sniffer methane concentrations showed dilution effects from ambient sampling, with median values of 83~ppm (IQR: 30--172~ppm) for Bonita and 126~ppm (IQR: 43--288~ppm) for Brittany---typically 100 to 1000 times lower than corresponding SCOUT values, reflecting the difference between direct ruminal sampling and ambient detection. Both animals spent similar total time under the feeding hood (17.4\% and 16.9\%), though Bonita ate fewer, longer meals while Brittany ate more frequently in shorter periods, consistent with their respective diets. The limited hood occupancy meant only approximately 17\% of sniffer measurements captured actual animal emissions. Table~\ref{tab:sniffer-signal-processing} reports sniffer signal characteristics.

\begin{table*}[!ht]
    \centering
    \small
    \caption{Sniffer-derived Methane Characteristics Summary}
    \resizebox{\textwidth}{!}{
        \begin{tabular}{lcccccc}
            \toprule
            \textbf{Cow ID} & \textbf{\% Time in} & \textbf{\#Events} & \textbf{Avg. Ambient} & \textbf{CH$_4$ Median} & \textbf{CH$_4$ IQR} & \textbf{\#CH$_4$ Peaks} \\
                            & \textbf{Feed Hood}  & \textbf{/day}     & \textbf{Drift (ppm)}  & \textbf{(ppm)}         & \textbf{(ppm)}      & \textbf{/day}           \\
            \midrule
            Bonita          & 17.4                & 19.5              & 315.9 $\pm$ 193.3     & 83.7                   & 30.1-171.9          & 12.1                    \\
            Brittany        & 16.9                & 27.9              & 61.0 $\pm$ 31.5       & 125.9                  & 42.6-287.9          & 16.4                    \\
            \bottomrule
        \end{tabular}
        \label{tab:sniffer-signal-processing}
    }
\end{table*}

\subsection{Cross-Platform Correspondence and Signal Characteristics}

Cross-platform correlation analysis revealed systematic scale-dependent relationships between SCOUT and sniffer measurements, with correlation strength increasing with analysis window duration. Short-duration windows (5~minutes) yielded weak correlations (r = $-0.077 \pm 0.235$) with only 47.5\% achieving statistical significance after BH-FDR correction. At 20-minute windows, correlations strengthened to r = $-0.330 \pm 0.138$ with 96\% significance.
The strongest correspondence emerged at 40-minute windows: r = $-0.564 \pm 0.007$ for signed regression coefficients, with 100\% of windows achieving statistical significance. This duration aligns with established ruminal fermentation and eructation cycles. The consistent negative correlation reflects the expected physiological coupling---SCOUT concentrations decrease during gas expulsion while sniffer concentrations increase due to methane accumulation in feeding enclosures---confirming that both systems detect the same underlying biological process. However, only a small fraction of \sens-detected events were mirrored in sniffer data, as the sniffer depends on cow positioning, eructation timing, and hood entry behavior. This moderate correspondence is consistent with the expectation that headspace concentrations and ambient sniffer readings reflect the same biology through fundamentally different measurement pathways.

The signal characteristics of the two systems reflect these different sampling environments. SCOUT, sampling directly from the sealed rumen headspace with an intrinsic response time under 1~second (Section~\ref{sec:signal_preprocess}), produces signals that track physiologically consistent patterns: stable elevated baselines during active fermentation, sharp concentration drops coinciding with eructation events, and gradual post-feeding rises consistent with fermentation kinetics. The sniffer system, while achieving response times under 3~seconds, operates in an open ambient environment where the measured signal reflects not only the animal's methane release but also head positioning, airflow disruptions, and intermittent hood presence, resulting in greater high-frequency variability.

\begin{figure*}[ht]
    \centering
    \includegraphics[width=\linewidth]{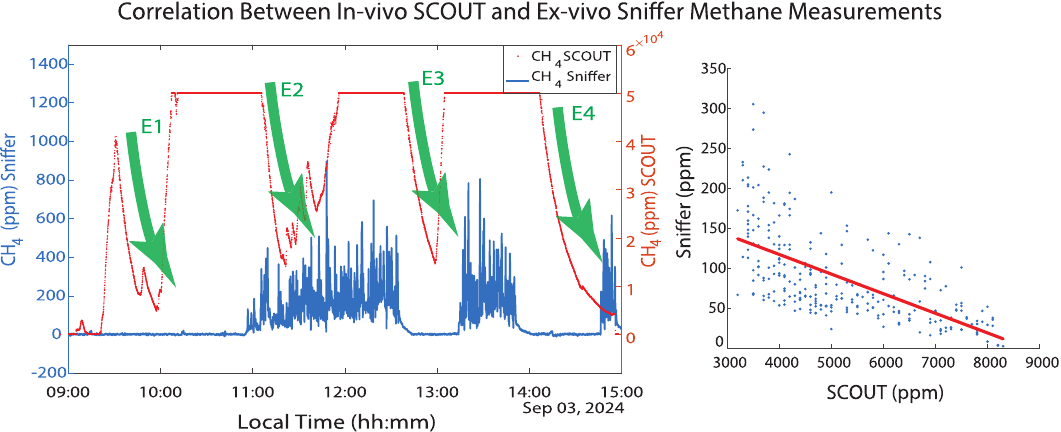}
    \caption{Temporal correlation between in-vivo SCOUT and ex-vivo sniffer measurements during eructation events. \textbf{Left:} Time series showing multiple eructation cycles (E1-E4) with SCOUT concentration drops (red, right axis) preceding corresponding sniffer peaks (blue, left axis). Green arrows highlight eructation events. \textbf{Right:} Scatter plot for a separate, single eruption event (chosen because SCOUT and Sniffer traces overlap over a 45-min window), with linear regression (R$^2$ = 0.39) between paired measurements.}
    \label{fig:cross-trend}
\end{figure*}

A representative comparison is shown in Figure~\ref{fig:cross-trend}, where both systems are plotted during overlapping eructation events. The temporal correspondence between SCOUT concentration changes and specific behavioral events (Figure~\ref{fig:video-behavior}) supports the biological plausibility of the \iv measurements. These differences in signal characteristics are expected given the two measurement contexts and do not indicate that either system is more accurate in absolute terms---rather, they access different aspects of the same methane production and release process. Prior work has shown that sniffer measurement fidelity deteriorates substantially under field conditions ($R^2$ from 0.97 in laboratory to 0.37 in commercial barns~\citep{wu2018uncertainty}), further contextualizing the observed cross-platform correspondence.

\subsection{Statistical Comparison of System Differences}

To evaluate whether dietary treatment produced consistent differences in measured methane concentrations across both systems, daily mean concentrations were computed for each system-animal-day combination ($N = 160$). A two-way ANOVA with Diet and Measurement System as fixed factors demonstrated a highly significant main effect for Diet ($F_{1,156} = 46.34$, $p < 0.001$; Table~\ref{tab:anova-results}), confirming that SCOUT distinguished between diets with statistical reliability comparable to the sniffer system. The significant main effect for Measurement System ($F_{1,156} = 42.42$, $p < 0.001$) reflects the expected difference in absolute concentration scales between the two sampling environments, while the significant interaction ($F_{1,156} = 42.16$, $p < 0.001$) is driven by this scalar difference. The strong Diet main effect indicates that, despite differences in absolute magnitude, SCOUT consistently tracks relative biological trends.

\begin{table}[ht]
    \centering
    \caption{Two-Way ANOVA on Daily Mean Methane Levels ($N=160$)}
    \begin{tabular}{lccccc}
        \toprule
        \textbf{Source}      & \textbf{Sum Sq.}      & \textbf{d.f.} & \textbf{Mean Sq.}     & \textbf{F-statistic} & \textbf{p-value} \\
        \midrule
        Diet                 & $7.55 \times 10^{14}$ & 1             & $7.55 \times 10^{14}$ & 46.34                & \textless 0.001  \\
        Measurement System   & $6.91 \times 10^{14}$ & 1             & $6.91 \times 10^{14}$ & 42.42                & \textless 0.001  \\
        Diet $\times$ System & $6.87 \times 10^{14}$ & 1             & $6.87 \times 10^{14}$ & 42.16                & \textless 0.001  \\
        Error                & $2.54 \times 10^{15}$ & 156           & $1.63 \times 10^{13}$ & --                   & --               \\
        \bottomrule
    \end{tabular}
    \label{tab:anova-results}
\end{table}

\subsection{Diurnal Patterns and Behavioral Correlations}
\label{sec:pattern_behavioral}

Analysis of diurnal emission patterns revealed consistent relationships between feeding schedules and methane production across animals and measurement systems. Peak concentration periods occurred 3--4~hours following major feeding events (0600 and 1500~hours), with maxima typically at 1200 and 1900~hours, aligning with established ruminal fermentation kinetics~\citep{blaise2018time}.

\subsubsection{Individual Differences}
\begin{figure*}[!bt]
    \centering
    \includegraphics[width=\linewidth]{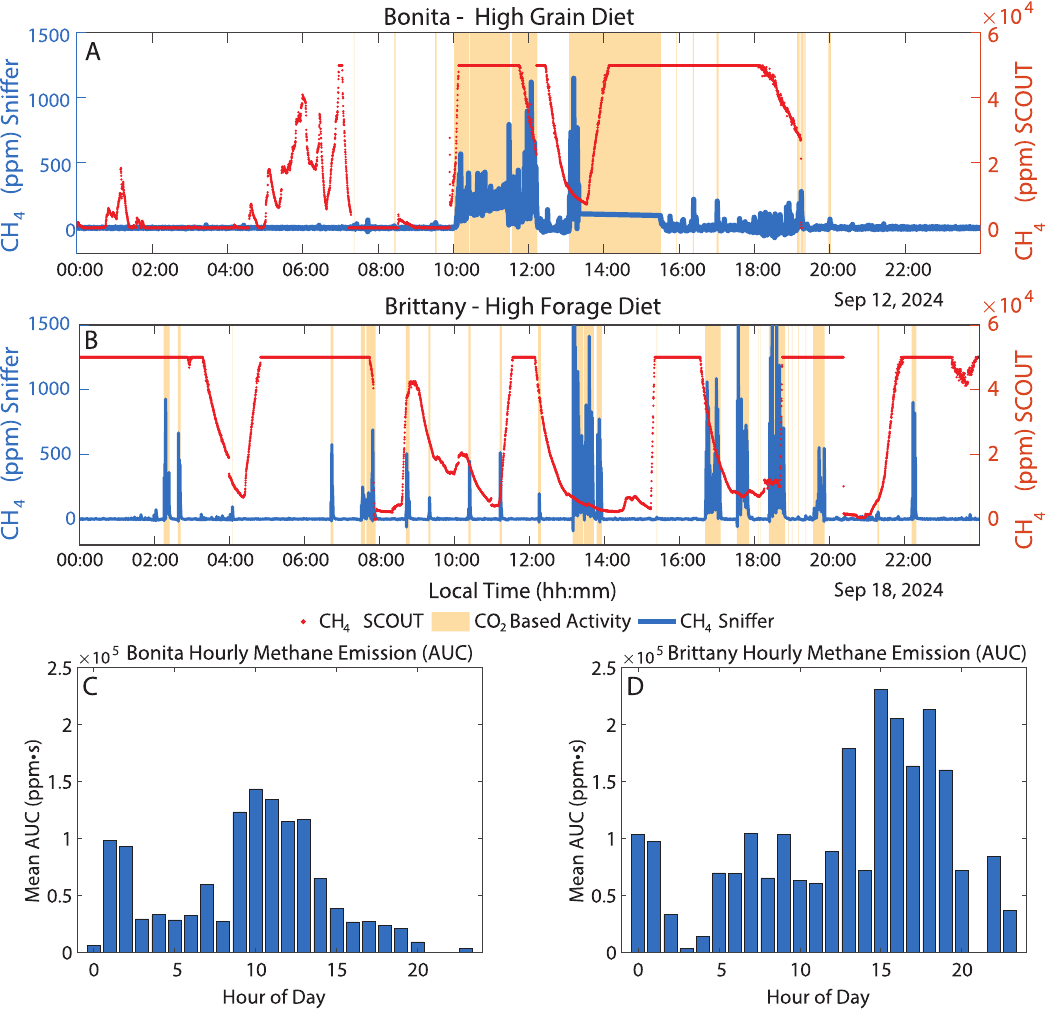}
    \caption{(A,B) Representative daily methane emission patterns for both cows under different dietary conditions. A sniffer \ch sensor reset was observed from 1~pm to 3~pm on Sep-12th, therefore the \ch data was removed from the dataset. (C,D) Hourly area-under-the-curve (AUC; ppm$\cdot$s) of Sniffer \ch emission for each hour of the corresponding 24-h period.}
    \label{fig:summary-both-cows}
\end{figure*}

Animal-specific concentration profiles emerged clearly despite similar housing and feeding protocols, as shown in Figure~\ref{fig:summary-both-cows}. Bonita displayed fewer but more intense peaks with consistent baselines, while Brittany exhibited higher overall variability with frequent moderate-intensity events. These differences persisted across multiple measurement days, indicating genuine biological variation rather than measurement artifacts, and were corroborated by sniffer data.

\subsubsection{Behavioral Triggers Quantified}Postural transitions provided particularly clear examples of behavior-concentration coupling (Figure~\ref{fig:video-behavior}). Standing-to-sitting transitions consistently triggered SCOUT concentration increases averaging 14.5k $\pm$ 11.3k ppm occurring within 15 $\pm$ 5 minutes of postural changes. This rapid response likely reflects gas redistribution or compression of rumen contents during postural adjustments.

\subsubsection{Feeding-Associated Temporal Dynamics}Feeding behavior correlations proved more complex than anticipated, with methane concentration changes typically lagging feed consumption by 15-45 minutes. This temporal delay suggests detectable emission changes reflect fermentation of newly consumed feed rather than mechanical gas release from existing rumen contents. The consistency of these temporal relationships across animals and days supports the biological relevance of observed patterns and validates the high-frequency monitoring utility for understanding methane production dynamics.

\begin{figure*}[ht]
    \centering
    \includegraphics[width=\linewidth]{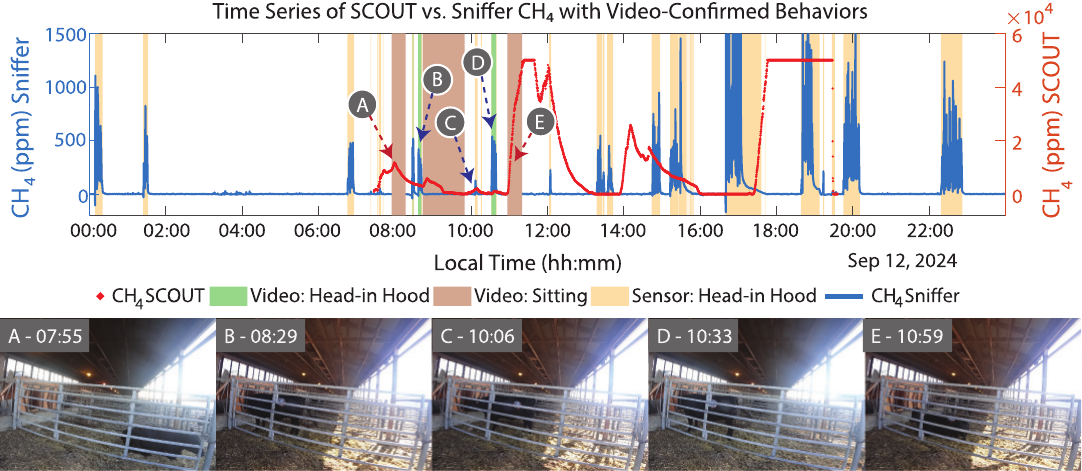}
    \caption{Time series of \ch from SCOUT (red, right axis) and Sniffer (blue, left axis) for Brittany on Sep-12, with behavior periods shaded (green: head-in-hood by video; orange: sitting by video; brown: head-in-hood by sniffer). Below are video stills A–E at their labeled timestamps, linked to the trace by dashed lines.}
    \label{fig:video-behavior}
\end{figure*}

Environmental sensitivity of the sniffer system became apparent when cows approached but did not fully enter feeding hoods, producing small but noticeable increases in both \co and \ch readings (Figure~\ref{fig:video-behavior}). This highlights potential influences of nearby animal activity on sniffer measurements in communal environments.

\section{Discussion}
\label{sec:discussion}
This study demonstrates the technical feasibility of continuous intraruminal methane monitoring through SCOUT's closed-loop gas recirculation design, addressing the engineering constraint that has limited direct ruminal methane measurement to discrete manual sampling events. The cross-platform correspondence with ambient sniffers ($r = -0.564$ at 40-minute windows) further confirms that SCOUT detects eructation events that propagate to the atmosphere, establishing a mechanistic link between the production-site and release-site signals. Together, these findings establish SCOUT as the first system capable of producing the continuous, high-resolution headspace concentration data that a concentration-to-flux model would require as its primary input.

Developing a validated concentration-to-flux model for SCOUT measurements requires characterising the relationship between headspace concentration time-series, eructation frequency and volume, rumen gas partitioning coefficients, and total emission flux as measured by a reference method such as respiration calorimetry. SCOUT provides the headspace concentration component of this model at temporal resolutions and signal magnitudes that prior systems could not achieve. The two to three orders of magnitude advantage in signal resolution over ambient methods means that concentration gradients associated with fermentation rate changes, eructation events, and behavioral perturbations are resolvable from measurement noise for the first time. This resolution is a prerequisite for fitting and validating the dynamic terms in any mechanistic production-to-emission model.

\subsection{Limitations}

The limitations of the current implementation warrant further consideration. The sensor's 50,000~ppm upper detection limit was exceeded during 35--41\% of measurements, predominantly during the post-feeding window of peak fermentation activity. The next-generation iterations will incorporate wider-range NDIR detectors, with commercially available alternatives extending to 100,000 ppm or beyond, to capture the full post-prandial fermentation response while balancing power consumption needs. Mechanically, 15 to 20\% of measurements were compromised by intake blockage or condensate accumulation, a problem addressable through refined water-trap geometry and hydrophobic membrane selection in subsequent prototypes. Further, for large-scale generalizations, the SCOUT system needs validation among larger, breed-diverse cohorts under a broader range of dietary treatments.  The final dataset derives from two animals under the finalized configuration, sufficient for proof-of-concept but requiring replication across larger, more diverse cohorts. Moreover, the cannula requirement restricts deployment to research settings where such access is already established, though this is consistent with SCOUT's intended role as a precision research instrument complementary to field-deployable methods, with miniaturisation toward a swallowable bolus configuration as a longer-term design trajectory.

\section{Conclusions}

SCOUT establishes that continuous, autonomous monitoring of ruminal methane concentrations is achievable through a cannula-mounted, closed-loop gas recirculation architecture that preserves anaerobic conditions throughout sustained \iv operation. Direct headspace sampling yielded concentrations orders of magnitude above concurrent ambient sniffer values, providing signal resolution sufficient to characterise behavioral coupling, diurnal fermentation patterns, and diet-induced differences that are fundamentally inaccessible through post-release measurement methods. Cross-platform comparison with ambient sniffers confirmed that SCOUT-detected concentration drops correspond to eructation events, establishing a mechanistic link between production-site and release-site signals at physiologically consistent timescales. The two-way ANOVA demonstrated reliable dietary discrimination across both measurement systems, corroborating the biological fidelity of the \iv concentration signal. Collectively, these results demonstrate that the rumen headspace contains continuous, biologically interpretable methane concentration signals that carry sufficient resolution to support the development and parametric validation of concentration-to-flux models. SCOUT provides the measurement infrastructure upon which such models, and the precision phenotyping, mitigation evaluation, and emission proxy calibration applications that follow from them, can be rigorously built.

\vspace{6pt}

\authorcontributions{Conceptualization, U.K. and R.M.V.; methodology, U.K.; software, Y.D. and U.K.; validation, Y.D., H.R.d.O. and U.K.; formal analysis, Y.D., H.R.d.O., R.M.V. and U.K.; resources, U.K., H.R.d.O., and R.M.V.; data curation, Y.D.; writing---original draft preparation, Y.D. and U.K.; writing---review and editing, U.K., H.R.d.O., and R.M.V; visualization, Y.D.; supervision, U.K.; project administration, U.K.; funding acquisition, U.K., H.R.d.O., and R.M.V. All authors have read and agreed to the published version of the manuscript.}

\funding{We acknowledge the support of Purdue Ag-Eng Seed funding in the completion of this work. We also acknowledge prior support of USDA grants 2019-67021-28990 and 2018-67007-28439 in the development of this work.}

\institutionalreview{The animal study protocol was approved by the Institutional Animal Care and Use Committee (IACUC protocol 0324002489).}

\informedconsent{Not applicable.}

\dataavailability{Data supporting the reported results will be made available upon reasonable request to the corresponding author.}

\acknowledgments{We acknowledge the support of Jihyo Park, Rieko Wilford, Yuanmeng Huang, as well as other undergraduate students who participated in the experiment and data collection activities.}

\conflictsofinterest{The authors declare no conflicts of interest.}

\reftitle{References}

\bibliography{example.bib}
\end{document}